\documentclass[conference,compsoc]{IEEEtran}

\usepackage{cite}
\usepackage{amsmath,amssymb,amsfonts}
\usepackage{graphicx}
\usepackage{textcomp}
\usepackage{xcolor}
\usepackage{diagbox}
\usepackage{hyperref}
\usepackage{subfigure}
\usepackage{multirow}
\usepackage{booktabs}
\usepackage{color}
\usepackage{adjustbox}
\usepackage{xurl}
\usepackage[ruled, vlined, linesnumbered, boxed]{algorithm2e}
\usepackage{booktabs} 
\usepackage{epstopdf}
\usepackage{makecell}
\usepackage[flushleft]{threeparttable}
\usepackage{listings}
\usepackage{xcolor}
\usepackage{balance}
\usepackage{tabu}
\usepackage{xspace}
\usepackage{multirow}
\usepackage{nimbusmononarrow}
\usepackage[T1]{fontenc}
\usepackage{textcomp,booktabs}
\usepackage{colortbl}
\usepackage{tabularray}
\usepackage{enumitem}
\usepackage{pifont}
\usepackage[all]{nowidow}
\usepackage{tcolorbox}
\usepackage{tikz}
\usepackage{caption}
\usepackage{filecontents}
\usepackage{marvosym}

\renewcommand{\paragraph}[1]{\vspace{5pt}\noindent{\bf{#1}.}}
\newcommand{\system}{\texttt{AVA}\xspace}
\newcommand{\xmark}{\ding{55}}%

\def\BibTeX{{\rm B\kern-.05em{\sc i\kern-.025em b}\kern-.08em
    T\kern-.1667em\lower.7ex\hbox{E}\kern-.125emX}}

\tcbuselibrary{breakable, skins}

\hypersetup{hidelinks}

\begin{document}

\title{AVA: Inconspicuous Attribute Variation-based Adversarial Attack bypassing DeepFake Detection}
	
\author{
    \IEEEauthorblockN{
        Xiangtao Meng\IEEEauthorrefmark{1},
        Li Wang\IEEEauthorrefmark{1},
        Shanqing Guo\IEEEauthorrefmark{1}\textsuperscript{(\Letter)},
        Lei Ju\IEEEauthorrefmark{1}, and
        Qingchuan Zhao\IEEEauthorrefmark{2}
    }
    \IEEEauthorblockA{\IEEEauthorrefmark{1}Shandong University\\\href{mailto:mengxiangtao@mail.sdu.edu.cn}{mengxiangtao@mail.sdu.edu.cn}, \href{mailto:wangli.tz@mail.sdu.edu.cn}{wangli.tz@mail.sdu.edu.cn}, \href{mailto:guoshanqing@sdu.edu.cn}{guoshanqing@sdu.edu.cn}, \href{mailto:julei@sdu.edu.cn}{julei@sdu.edu.cn}}
    \IEEEauthorblockA{\IEEEauthorrefmark{2}City University of Hong Kong\\ \href{mailto:qizhao@cityu.edu.hk}{qizhao@cityu.edu.hk}}
}

\maketitle

\begin{abstract}
While DeepFake applications are becoming popular in recent years, their abuses pose a serious privacy threat.
%
Unfortunately, most related detection algorithms to mitigate the abuse issues are inherently vulnerable to adversarial attacks because they are built atop DNN-based classification models, and the literature has demonstrated that they could be bypassed by introducing pixel-level perturbations.
%
%
Though corresponding mitigation has been proposed, we have identified a new attribute-variation-based adversarial attack (\system) that perturbs the latent space via a 
combination of Gaussian prior and semantic discriminator to bypass such mitigation.
It perturbs the semantics in the attribute space of DeepFake images, which are inconspicuous to human beings (e.g., \textit{mouth open}) but can result in substantial differences in DeepFake detection. 
%
%
%
%
We evaluate our proposed \system attack on nine state-of-the-art DeepFake detection algorithms and applications. 
The empirical results demonstrate that \system attack defeats the state-of-the-art black box attacks against DeepFake detectors and achieves more than a 95\% success rate on two commercial DeepFake detectors.
Moreover, our human study indicates that \system-generated DeepFake images are often imperceptible to humans, which presents huge security and privacy concerns. 
%
%
\end{abstract}

\section{Introduction}
DeepFake, which leverages advanced artificial intelligence expertise to digitally generate, manipulate, or replace one person's face without being trivially perceived by normal users, 
has been adopted in many popular applications that are convenient for people's lives, such as creating stylized images~\cite{li2017generative}, educating through the reanimation of historical figures~\cite{lee2019deepfake}, and virtually trying on clothes while shopping~\cite{virtuallytry}. 
However, this technique has also been abused for malicious behaviors, such as violating one person's privacy and reputation that 
often targeted political leaders, actresses, comedians, and entertainers and had their faces weaved into porn videos~\cite{hasan2019combating}.
It has been reported that 96\% of existing DeepFake videos online are pornography aimed at women around the world, with more than 134 million views~\cite{threats}. 
Other DeepFake abuses include internet bullying, fake video evidence in courts, terrorist propaganda, blackmail, market manipulation, and fake news~\cite{maras2019determining,westerlund2019emergence}.
%
%

%
There are tremendous efforts that have been devoted to detecting DeepFake images to suppress its related abuses~\cite{wang2020cnn, nguyen2019use, dang2020detection, liu2020global, qian2020thinking}; unfortunately, most of them are inherently vulnerable to adversarial attacks.
These detection algorithms are mostly designed based on the observation that the pixel space's distribution of DeepFake images differs from that of non-DeepFake images, though DeepFake images are realistic enough to be imperceptible by humans.
%
%
%
Accordingly, these algorithms utilize deep neural networks (DNN) to extract the pixel space's distribution of images for detection and achieve a high accuracy rate on related benchmarks.
%
%
However, the weak robustness of DNN makes these detections subject to adversarial attacks.
%
For example, carefully crafted pixel-level perturbation could reduce the differences between the pixel space's distribution of a DeepFake image and non-DeepFake images learned by DeepFake detections, making it harder to be detected~\cite{gandhi2020adversarial,carlini2020evading,neekhara2021adversarial,hussain2021adversarial,liao2021imperceptible,fan2021deepfake}.

Although there have been proposed solutions (e.g., ~\cite{pang2018towards}) to mitigate such an inherent vulnerability, this mitigation family is limited at the pixel level, making it fundamentally insufficient for DeepFake detection.
Interestingly, we found that a DeepFake image could be generated not only by manipulating the pixels but also the attributes, such as mouth, hairstyle, and eyebrow, and their semantics.
For example, a DeepFake image could make one's mouth (the attribute) open more widely (the semantics) and, similarly, a different hairstyle or eyebrow shape.
Also, surprisingly, our preliminary study found that attribute-level attacks could be more powerful than pixel-level attacks.
%
First, introducing attribute-level perturbations to change the attributes' semantics has limited effects on the associated pixel space's distribution because it has bypassed the state-of-the-art of both academic white-box detectors and commercial black-box detectors that are built atop pixel-level variations. 
Second, unlike most pixel-level adversarial attacks that assume full access to the victim detectors' model, our attribute-variation-based attack has no such assumption and can totally bypass black-box detectors.
In addition, we also found this new attack is significantly more effective than related black-box attacks (i.e., Hussain \textit{et al.}~\cite{hussain2021adversarial} and Fakepolisher~\cite{huang2020fakepolisher}).
Given that such an attribute-level perturbation has not been well-studied in DeepFake security, we propose to explore the feasibility of this new attribute-variation-based adversarial attack (\system attack) in perturbing the latent space to bypass the state-of-the-art DeepFake detections. 
Specifically, 
it first inverts a DeepFake image into the GAN latent space to represent its multiple features for convenient manipulation,
and then searches for the optimal semantic attributes to apply perturbations.
However, it is non-trivial to achieve these two objectives.
Specifically, the DeepFake image may be inverted to the low-density portions of the GAN latent space, resulting in abnormal images after attribute manipulation. 
For example, if the generated face's mouth opens too exaggeratedly, this DeepFake image will obviously appear abnormal because it is unlikely to be a normal human face.
Additionally, the highly irregular distribution of the attribute space makes it challenging to constrain semantic perturbations, always remaining within the reasonable attribute space.
Though setting a threshold appears to be plausible, it is challenging to establish the boundaries of the attribute space properly because a smaller threshold reduces the space of semantic perturbations, making it easier to be detected, and a larger one could make perturbations exceed the normal space, making generated faces abnormal.
%
\looseness=-1
%


We leverage two insights to address the challenges in launching our proposed \system attacks.
Specifically, our first insight to address the challenge of low-density portions in the latent space is that such low-density portions could be optimized to be high-density portions.
We transform the inverted latent code in the latent space into an auxiliary space that follows a multi-variate Gaussian distribution. 
Thus, we can fine-tune the latent code towards high-density portions of the latent space using a Gaussian prior.
Additionally, our second insight to address the challenge of constraints on semantic perturbations is that a semantic discriminator could approximately model the distribution of the attribute space.
The semantic discriminator is trained adversarially to learn the semantic distribution of normal samples with various attributes. 
Alongside the increase of the training dataset, the distribution of the attribute space could be modeled. 

We have implemented a prototype to demonstrate the feasibility of our proposed attack and evaluate it on nine state-of-the-art DeepFake detection algorithms, including both white-box ones (e.g., CNNDetection~\cite{wang2020cnn}, FFD~\cite{dang2020detection}, and F$^3$-Net~\cite{frank2020leveraging}) and black-box ones (e.g., Baidu~\cite{baidu} and Tencent~\cite{tencent}). 
Our empirical results demonstrate that our proposed \system attack can bypass all state-of-the-art detection algorithms involved in this study alongside a higher visual performance than three pixel-noise-based attacks.
In particular, our attack success rate on two commercial DeepFake detectors reaches over $95\%$, which presents a huge security risk to various downstream applications using such detectors (e.g., embedding associated APIs).
%
The empirical results on two benchmark datasets show that our proposed \system attack applies to all types of DeepFakes, not only the GAN-based DeepFake.
Besides, the human study results demonstrate that \system is often imperceptible to humans.
Moreover, our empirical results also demonstrate that the \system attack has a more effective penetrating capability to the state-of-the-art defense methods than three pixel-level-perturbation-based attacks. 

\paragraph{Responsible disclosure}
We have reported our results to the corresponding vendors and organizations and have obtained their acknowledgments. 
%
We are now and will continue to engage with them to offer help with our best efforts. 

Our main contributions are summarized as follows:
\vspace{-3pt}
\begin{itemize}[leftmargin=*]
\item \textbf{New attack.} we have identified a new attribute-variation-based adversarial attack (\system) that perturbs the latent space via a combination of Gaussian prior and semantic discriminator to bypass the state-of-the-art DeepFake detectors, which may impact billions of users' privacy.

\item \textbf{Novel technique.} 
We utilize a Gaussian prior in an auxiliary space to optimize the DeepFake image inverted in low-density portions to high-density portions.
Additionally, we train a semantic discriminator to approximately model the distribution of the attribute space, thus constraining the semantic perturbations in a more proper boundary than a simple threshold.

\item \textbf{Empirical evaluation.} The empirical results demonstrate that our proposed attack can bypass the nine state-of-the-art DeepFake detection algorithms involved in this study and defeat the prominent state-of-the-art black box attacks against DeepFake detectors.

\end{itemize}

\section{Related Work and Threat Model}

\subsection{DeepFake: Key Techniques}
DeepFake is a facial image synthesis technology based on deep learning, which can generate or manipulate facial appearance without being perceived by human beings.
%
Existing DeepFakes can be mainly classified into four categories based on their function: Entire face synthesis~\cite{karras2019style, karras2020analyzing, karras2017progressive}, which aims to generate non-existent fake face image; Attribute manipulation~\cite{choi2020stargan,li2021image,wu2021stylespace}, which aims to modify facial attributes of a real image; Identity swap~\cite{nirkin2019fsgan,DBLP:conf/mm/ChenCNG20,li2020advancing}, which aims to replace the identity of source image; Expression swap~\cite{zhang2021facial, Wang_2021_CVPR}, which aims to replace the expression of source image. 
The development of DeepFakes has promoted a wide range of benign and malicious applications, \textit{e.g.}, visual effect assistance in movies and misinformation generation by faking famous persons.

\vspace{-0.1cm}
\subsection{DeepFake Detection}
DeepFake detections rely on detecting manipulation artifacts resulting from Deepfake generation to classify images as ``Real`` or ``Fake``. Thus far, these detections are summarized as follows:

\paragraph{Spatial-domain based detection} These detections observe visible or invisible artifacts on the spatial domain of images to distinguish real and fake. 
They identify these artifacts based on different network architectures, processing strategies, and extracted features.
For example, VA-MLP\cite{matern2019exploiting} manually captures relatively simple visual artifacts in DeepFake images, such as eye color, missing reflections, and missing details in the eye and teeth regions. 
CNNDetection\cite{wang2020cnn} incorporates various data augmentation strategies into the training process to improve the robustness of DeepFake detections.
Patch-forensics~\cite{chai2020makes} utilizes shallow networks with limited receptive fields to focus on small patches of the DeepFake image to improve detection accuracy.

\paragraph{Frequency-domain based detection} These detections first convert the image to the frequency domain and then extract features in the frequency domain. They normally mine distinctive features from the image spectra via statistical analysis or machine learning. For example, F$^3$-Net~\cite{frank2020leveraging} determined that there were Fourier spectrum discrepancies between CNN-generated images and real images which could be efficiently captured by a shallow machine learning classifier. Qian \textit{et al.}~\cite{qian2020thinking} mine frequency-aware clues from both frequency-aware decomposed image components and local frequency statistics.

\subsection{Attacks against DeepFake Detection}


DeepFake detections play a vital role in curbing the abuse of DeepFakes. 
Therefore, increasing efforts have been devoted to exploring the vulnerabilities of existing DeepFake detections to ensure their reliability. 
These works are summarized as follows:

\paragraph{Reconstruction-based attacks}
The reconstruction-based attacks~\cite{huang2020fakepolisher, 9133490} adopt the idea of GAN~\cite{goodfellow2014generative} to train a generative model which is designed to reconstruct the DeepFake images by projecting them onto the manifold subspace learned from real images to mislead DeepFake detection methods. 
Fakepolisher~\cite{huang2020fakepolisher} is the most representative work.
However, it only aims to reduce artifacts in DeepFake images by projecting them to the true data distribution.
Thanks to the detection method can not accurately learn the true data distribution, most adversarial examples may be away from the distribution area of real images learned by DeepFake detections, making them can not successfully bypass the unknown DeepFake detections.
%

\paragraph{Iterative adversarial attacks}
Several works~\cite{gandhi2020adversarial,carlini2020evading,neekhara2021adversarial,hussain2021adversarial,liao2021imperceptible,fan2021deepfake} have explored the vulnerability of DeepFake detections using several pixel-level-perturbation-based iterative adversarial attacks, including the Fast Gradient Sign Method (FGSM)~\cite{goodfellow2014explaining}, Carlini and Wagner l2-norm Attack~\cite{carlini2017towards}, and Projected Gradient Descent (PGD)~\cite{madry2017towards}. They all show that these iterative adversarial attacks can still be applied to fool DeepFake detections. 
%
However, due to the detection method not accurately learning the true data distribution, most adversarial examples generated by these attacks may be off the true data distribution though they bypass DeepFake detections.
Therefore, corresponding mitigation approaches such as adversarial training~\cite{madry2017towards}, can easily make these attacks invalid.
Recently, Li \textit{et al.}~\cite{li2021exploring} proposed to search for adversarial perturbation on the latent space of StyleGAN to generate enhanced fake face images that can bypass DeepFake detections. 
%
However, due to the complexity and high-dimensionality of the latent space, even a small perturbation added may lead to substantial and uncontrollable variations in identity features, thereby reducing the searchable perturbation space and significantly enhancing the likelihood of human perception of the generated adversarial examples.
\vspace{-0.1cm}
\subsection{Threat Model}




Here, we state assumptions for both the adversary bypassing DeepFake detectors and the victim DeepFake detectors detecting DeepFake images.

\paragraph{Adversary}
We assume the adversary has access to a released DeepFake tool and moderate computing resources (e.g., a laptop). The adversary wants to create DeepFake images using the DeepFake tool and publish them to social network platforms (e.g., Facebook~\cite{facebook}) with a large user base for various malicious purposes.
For example, an adversary may wish to defame the opponent candidate in a political campaign.
However, these social network platforms may deploy DeepFake detectors to detect images uploaded by users, which prevents DeepFake images generated by the adversary from spreading. Moreover, users who receive these DeepFake images may deploy DeepFake detectors locally to prevent being fooled by them.
Therefore, the goal of the adversary is to purposefully perturb the human-imperceptible semantics in the attribute space of generated DeepFake images to bypass these DeepFake detectors.

\paragraph{Victim DeepFake detectors}
The victim DeepFake detectors' goal is to detect DeepFake images to suppress their abuses. A DeeFake detector could be developed by an industry-leading vendor for commercial use, e.g., Tencent~\cite{tencent} or Microsoft Video Authenticator~\cite{MicrosoftVideoAuthenticator}, or an individual researcher for academic research, e.g., CNNDetection~\cite{wang2020cnn}.

We assume both white-box and black-box attack scenarios where an adversary has complete access to the victim model, including the architecture, parameters, and the output of the victim model in the white-box scenario, but the adversary can only obtain the output without any other information about the model in the black-box scenario. 
These assumptions are reasonable for the following reasons.
First, most commercial DeepFake detectors are embedded in social platforms to prevent the spread of unsafe DeepFake images.
These commercial DeepFake detectors are kept confidential (i.e., black-box) without disclosing their model architecture and parameters but leaving ways to obtain results (e.g., via APIs).
Second, most academic detectors are open-sourced (i.e., white-box) by individual researchers, which are usually used for free by individuals or organizations to detect suspicious images (Section~\ref{section:evaluation}).
Moreover, we also assume the developers of the victim DeepFake detector are aware of our attack and can deploy adaptive countermeasures (Section~\ref{section:countermeasures}).

\section{Approach }
\begin{figure*}[htp]
    \centering
    \includegraphics[width=18cm]{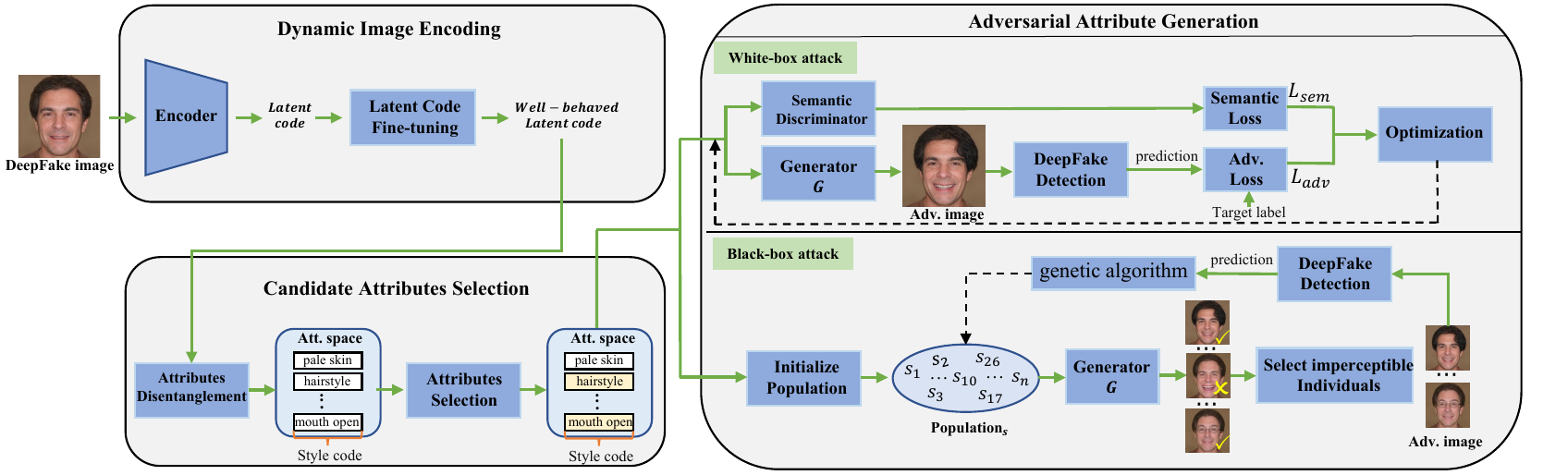}
    \caption{The overall architecture of our proposed \system attack.}
    \label{fig:fig1}
    \vspace{-0.2cm}
\end{figure*}

In this section, we have identified an attribute-variation-based adversarial attack (\system) with a novel technique that hasn't been tried before, i.e., the combination of Gaussian prior and semantic discriminator, to bypass DeepFake detectors. Specifically, we will first present the overall of our proposed \system attack architecture, then we elaborate on the details of the design and implementation of our attack.

\subsection{Overview}
Figure~\ref{fig:fig1} presents an overall architecture of our proposed \system attack. The \system attack aims to adversarially edit inconspicuous attributes of the DeepFake image to bypass DeepFake detection, which is based on the face manipulation approach, StyleGAN2~\cite{karras2020analyzing}. 
Our attack consists of three modules: \textit{Image Embedding}, \textit{Candidate Attributes Selection}, and \textit{Adversarial Attribute Generation}. 
First, to make the generated adversarial examples closer to the true data distribution, we embed the DeepFake image into the GAN latent space via an encoder trained on real images.
Then we shift the embedded latent code to a more well-behaved latent code embedding in high-density portions of the GAN latent space via corresponding constraints. This can prevent the generation of abnormal faces after attribute manipulation. 
Next, the well-behaved latent code is sent to the candidate attributes selection module. 
In this module, we first decouple the latent code into a space in which attributes are not entangled with each other via an attribute disentanglement operation, resulting in a disentangled style code. 
Then, we mark the region corresponding to the candidate attributes on the style code. 
Next, the marked style code is sent to the adversarial attribute generation module. 
In the adversarial attribute generation module, we design the attack algorithm in white-box and black-box scenarios respectively. 
In the white-box scenario, the style code is fed to the semantic discriminator to obtain the semantic loss, denoted by $L_{sem}$. This can constrain the semantic perturbations always remaining within the reasonable attribute space. Additionally, based on the style code, the generator $G$ of StyleGAN~\cite{karras2020analyzing} generates an adversarial example which is fed to the DeepFake detection to get its prediction. Then, it computes the adversarial loss between the prediction and the target label, denoted by $L_{adv}$. $L_{sem}$ and $L_{adv}$ work together to optimize the style code until successfully bypass DeepFake detection or reach the maximal iteration. 
In the Black-box attack, we first randomly sample the input style code in candidate attribute spaces to initialize the first population, denoted by Population$_s$. Then generator $G$ generates a set of samples based on these style codes in Population$_s$. Next, we will select inconspicuous individuals to form our adversarial example sets which are fed to the DeepFake detection to get their predictions. Subsequently, we utilize the genetic algorithm~\cite{bhandari1996genetic} to generate a new population to continue until successfully bypass DeepFake detection or reach the maximal iteration.

\subsection{Dynamic Image Encoding Module}\label{section:ganinversion}
To change the attributes' semantics, encoding the DeepFake image input into the latent space of StyleGAN is an efficient solution.
However, since our approach aims at all types of DeepFakes, the unknown DeepFake image may be embedded in the low-density portions of the latent space by the encoder, resulting in the generation of abnormal images or with unstable semantics~\cite{zhu2020improved}. 
Therefore, we propose a dynamic image encoding module to obtain a DeepFake image's well-behaved latent code.
The well-behaved latent code refers to the latent code embedded in the high-density portions of the latent space.
This module consists of the following two steps:
Encoder, which aims to encode a DeepFake image to a latent code within the latent space; 
Latent Code Fine-tuning, which aims to fine-tune the latent code embedded in the low-density portions of the latent space to a well-behaved latent code. 
%
%

\paragraph{Encoder}
In this step, we utilize an encoder trained on real images to map the DeepFake image to a latent code in the latent space of StyleGAN~\cite{karras2020analyzing}.
%
The StyleGAN involves several latent spaces. 
Firstly, it starts with the initial latent space, denoted as $Z$, which is typically normally distributed.
Then the random noise vectors $z \in Z$ are transformed into an intermediate latent space $W$ via a sequence of fully connected layers.
Subsequently, different intermediate latent codes $w$ corresponding to each layer of the generator compose the $W^+$ space. 
Finally, the affine transformation operation maps the latent code in $W^+$ space to $S$ space.
Previous works, such as \cite{abdal2019image2stylegan,wulff2020improving}, have shown that the expressiveness of the $W$ and $Z$ latent space of StyleGAN is limited, in that not every image can be accurately mapped into them. 
Besides, Wu \textit{et al.} point out that the $S$ space provides high visual accuracy of the reconstruction compared with $W^+$ space, but the worse manipulation naturalness. 
To achieve a satisfactory compromise between reconstruction accuracy and manipulation naturalness, we adopt the e4e encoder~\cite{tov2021designing} to embed the DeepFake image into the $W^+$ space.

\paragraph{Latent Code Fine-tuning}
In this step, to prevent input from being embedded in low-density portions of StyleGAN's latent space, we design a regularization term to fine-tune the latent code to a well-behaved latent code under the premise of pixel-wise similarity.
However, due to the distributions of $W$ and $W^+$ space being highly irregular, it is challenging to fine-tune the latent code towards high-density portions of $W^+$ space.
Fortunately, it could be found that the irregular distribution is due to the very last processing operation before computing latent code $w$, i.e.\ Leaky ReLU with a fixed negative slope of 0.2 \cite{wulff2020improving,xu2020d}. The distribution before the leaky ReLU operation follows a multi-variate Gaussian distribution, denoted $P$ space \cite{wulff2020improving, zhu2020improved}. Therefore, we could fine-tune the latent code with $P$ space following a multi-variate Gaussian distribution. 

To constrain the latent code $w^+$ via the $P$ space with a regular distribution, we should first transform $w^+$ to $w$. The relationship between $w$ and $w^+$ is as follows:
\begin{equation}
w^{+}=\left \{ w_{i}  \right \}_{i=1}^{18}   
\end{equation}

Then the latent code $w$ in $W$ space is transformed into the $P$ space by inverting the last LeakyReLU layer in the StyleGAN mapping network as follows:
\begin{equation}
p = LeakyReLU_{5.0}(w)
\end{equation}

where $w \in {W}$ and $p \in {P}$. We take a large number of Gaussian samples from the $Z$ space and feed them to the mapping network $f$ and calculate the empirical covariance matrix $\Sigma$ and mean $\mu$ of the distribution in $P$ space.

Through the above transform, we can obtain the space following a multi-variate Gaussian distribution corresponding to the $W^+$ space, denoted $P^+$ space.
\begin{equation}
p^{+}=\left \{ p_{i}  \right \}_{i=1}^{18}=\left \{ LeakyReLU_{5.0}(w_{i})  \right \}_{i=1}^{18}   
\end{equation}

Then, we convert the Gaussian prior of $P^+$ space to an energy term as the regularization term for latent code fine-tuning as follows:

\begin{equation}
\left ( p^{+}-\mu^{+}    \right ) \Sigma ^{+^{-1} } \left (  p^{+}-\mu^{+} \right ) 
\end{equation}

where $\Sigma ^{+} = \mathrm{I} _{18}\otimes \Sigma$ and $\mu^{+}=\left \{ \mu_i  \right \}_{i=1}^{18}$. 
Thanks to the network architecture of StyleGAN consists of 18 layers, and each layer corresponds to a $P$ space, we choose $\mu$ to represent the mean of the distribution in each $P$ space, \textit{i.e.}, $\left \{ \mu_i = \mu \right \}_{i=1}^{18}$.

To guarantee the pixel-wise similarity between the original image and the reconstructed image during optimization, we adopt the LPIPS \cite{zhang2018unreasonable} metric to calculate the reconstruction loss, denoted $L_{lpips}$.

Thus, we can optimize the latent code $w+$ by the following equation:
\begin{align}
w_{i+1}^{+}= & arg\min_{w_{i}^{+}\in W^{+}} L_{lpips}\left ( x, G\left ( w_{i}^{+} \right )  \right ) \notag \\ + & \lambda \left ( p_{i}^{+}-\mu^{+}    \right ) \Sigma ^{+^{-1} } \left (  p_{i}^{+}-\mu^{+} \right )  
\end{align}
where $x$ is the DeepFake image input, $G\left ( \cdot  \right ) $ is the StyleGAN generator, and $\lambda$ determines the weight of the regularization term during latent optimization. 
The GAN inversion module can be formally described by Algorithm \ref{alg1}.

\SetKwRepeat{Do}{do}{while}%
\begin{algorithm}\small
\caption{GAN Inversion Function}
\label{alg1}
\KwIn{ A DeepFake image $x$; hyperparameter $\lambda$; the number of iterations $n$; a pre-trained generator $G(\cdot)$; a pre-trained encoder $E(\cdot)$; the empirical covariance matrix $\Sigma^+$ and mean $\mu^+$ of the distribution in $P^+$ space.}
\SetKwFunction{Inversion}{Inversion}
\SetKwProg{Fn}{Function}{:}{}
\Fn{\Inversion{$x$, $\lambda$, $n$, $G$, $E$, $\Sigma^+$, $\mu^+$}}{
        $w_0^+=E(x)$, $w^+=w_0^+$\;
        \For{j=0:n}{
            $p_{j}^{+} = LeakyReLU_{5.0}(w_j^+)$ \;
            $w_{j+1}^{+}= \mathop{\arg\min}_{w_{j}^{+}\in W^{+}} \ L_{lpips}\left ( x, G\left ( w_{j}^{+} \right )  \right )  + \lambda \left ( p_{j}^{+}-\mu^{+}    \right ) \Sigma ^{+^{-1} } \left (  p_{j}^{+}-\mu^{+} \right )$ \;
            $w^+=w_{j+1}^+$\;
            $j = j + 1$\;
        }
        \KwRet $w^+$\;
  }
\end{algorithm}

\subsection{Candidate Attributes Selection Module}\label{section:attributecombination}
In the previous module, we obtained the well-behaved latent code in StyleGAN's latent space.
However, features and attributes in the latent space are often entangled, \textit{i.e.}, making small changes to the latent code may affect multiple attributes simultaneously, even causing large variations in identity. 
Therefore, in this module, we first decouple the attributes of the well-behaved latent code and then select candidate attributes in the decoupled attribute set.
%

\paragraph{Attributes Disentanglement}
In this step, we aim to decouple the attributes of the well-behaved latent code.
Wu \textit{et al.} point out that $S$ space is better suited for fine-grained and disentangled visual manipulation, compared to other latent spaces of StyleGAN, thus we first transform the well-behaved latent code $w^+$ to $S$ space to obtain the style code $s$ using affine transformation. 
The style code consists of 9088 StyleSpace channels, and recent research~\cite{wu2021stylespace} suggests that some channels in the style code can control the variation of specific attributes, such as mouth open.
Specifically, research \cite{wu2021stylespace} proposes to obtain channels that control the specific attributes in style code by computing the gradient maps of generated images with respect to different style parameters and measuring their overlap with specific attribute regions.
Thus, these channels in the style code compose a decoupled attribute set, denoted $S^*$.

\paragraph{Attributes Selection}
Having obtained the disentangled attribute set $S^*$, in this step, we aim to select candidate attributes for \system.
Specifically, we define the state of these channels in the style code as editable or non-editable. 
The editable state means that the value of the channel can be changed (i.e., the attribute controlled by the channel can be altered) while the non-editable is the opposite. 
Note that the state of those non-disentangled channels is set non-editable to prevent generating images far from the true distribution or largely changing identity features.
Here, we define a vector $\vec{s}$ to represent the state of each channel in the style code, where the editable state is set to 1 and the non-editable state is set to 0.
%
%
%
Therefore, we can select candidate attributes set $S^c$ via altering the vector $\vec{s}$.

\subsection{Adversarial Attribute Generation Module}\label{section:adversarialattribute}

In real attack scenarios, the adversary possesses varying levels of knowledge and information available about the victim model. 
In this module, we consider the following two traditional scenarios to characterize an adversary's background knowledge: White-box scenario, where the adversary can completely access to the DeepFake detector, including the architecture and parameters; Black-box scenario, where the adversary has no knowledge of the DeepFake detector except output.
We have designed attack methods for each of the two scenarios mentioned above, as there is currently no unified method that performs well in both.

\subsubsection{White-box scenario} 
In the white-box scenario, the adversary can completely access the DeepFake detector. 
Therefore, we can perform our attack under the guidance of the victim model's gradients with respect to the candidate attributes. 
The gradients provide information about how various attribute variations will affect the model's output.
However, two requirements must be met to achieve our attack goals: Move the adversarial examples closer to true data distribution and move the adversarial examples closer to the data distribution of real images learned by DeepFake detections.
Therefore, we establish the following two loss functions to meet the above requirements.
\begin{itemize}[leftmargin=*]
\item \textbf{Semantic loss.} 
To guarantee the generated adversarial examples are closer to the true data distribution, we should constrain the semantic perturbations always remaining within the reasonable attribute space.
Unfortunately, the highly irregular distribution of the attribute space makes it challenging. 
Though setting a threshold appears to be plausible, it has difficulty in properly describing the boundaries of the attribute space, which may result in two extreme cases. 
%
%
First, if the threshold is too small, the attribute space in which we search for semantic perturbations is correspondingly reduced, reducing the probability of finding semantic perturbations that bypass DeepFake detection.
Second, if the threshold is too large, the semantic perturbations we get may exceed the attribute space, resulting in abnormal generated faces.
Inspired by \cite{tov2021designing}, we train a semantic discriminator to approximately model the distribution of the attribute space, denoted $D_s$. 
The semantic discriminator is trained in an adversarial manner to discriminate between real samples from the S space (generated by StyleGAN's mapping function and affine transformation) and the encoder's learned poor latent codes.
In this way, the semantic discriminator can learn the semantic distribution of normal samples with various attributes. As the training data set increases, it can approximately model the distribution of the attribute space.
The loss used in training the semantic discriminator is as follows:

\begin{align}\label{eq:7}
    L_{adv}^{D}=- & \underset{s\sim S}{E}\left [ \log{D_{s}\left ( s \right ) }  \right ]-  \underset{x\sim pX}{E}\left [ \log{\left ( 1-D_{s}\left ( E\left ( x \right )_i  \right )   \right ) }  \right ]\notag \\ +  &  \frac{\gamma }{2}\underset{s\sim S}{E}\left [ \left \| \bigtriangledown_{s}D_{s}\left ( s \right )    \right \|_2^2  \right ]     
\end{align}
\begin{equation}
\begin{aligned}\label{eq:8}
   & L_{adv}^{E}=-\underset{x\sim pX }{E}\left [ \log {D_{s}\left ( E\left ( x \right )_{i}  \right ) }  \right ]    
\end{aligned}
\end{equation}

Based on the semantic discriminator, we can obtain the semantic loss as follows:
\begin{equation}
L_{sem} =  D_s\left ( s  \right )
\end{equation}

\item \textbf{Adversarial loss.}
To successfully bypass DeepFake detections, we should search the semantic perturbation that can move the DeepFake image to the distribution area of real images learned by DeepFake detections in the attribute space. 
Therefore, we design an adversarial loss, $L_{adv}$ as follows to minimize the difference between the model's output and the target label.
\begin{equation}
L_{adv} =  loss\_func\left ( D ( G ( s ) ), t  \right )
\end{equation}
where $loss\_func()$ is the loss function used by the victim DeepFake detection, $D$ is the victim DeepFake detection, $G$ is the StyleGAN generator,  $s$ is the style code of input, and $t$ is the target label.
\end{itemize}

To sum up, the adversarial attributes can be updated as follows until successfully bypassing the DeepFake detections or reaching the maximum iteration.
\begin{equation}
s_{i+1} = s_i + variation ( \vec{s} , \nabla_{s\in S^c} (L_{adv} + \alpha L_{sem} ) )
\end{equation}

where $\alpha$ is the trade-off hyperparameter, $variation()$ is the function to calculate the direction and magnitude of the selected candidate attributes' manipulation.

\SetKwRepeat{Do}{do}{while}%
\begin{algorithm}\footnotesize
\caption{White-box Adversarial Optimization Function}
\label{alg2}
\KwIn{ An DeepFake image $x$; style code $s$; trade-off hyperparameters $\alpha$; size of population $n$; target label $t$; maximum iterations $max$; the DeepFake detector $D(\cdot)$.}

\SetKwFunction{StateChange}{StateChange}
\SetKwFunction{Adam}{Adam}
\SetKwFunction{GetPreds}{GetPreds}
\SetKwFunction{FindElite}{FindElite}
\SetKwFunction{InitPopulation}{InitPopulation}
\SetKwFunction{White}{White-box Optimization}
\SetKwProg{Fn}{Function}{:}{}

\Fn{\White{$x, s, \alpha, n, t, max, D$}}{
    $s_0=$ \StateChange{$s$}\;
    $i = 0$; $pred \neq t$\;
    \While{$pred \neq t$ and $i < max$}{
        $\hat{x} =G(s_i)$\;
        $Loss = loss\_func\left ( D ( G ( s ) ), t  \right ) + \alpha D_s\left ( s  \right )$\;
        $s_{i+1}= s_i + variation ( \vec{s} , \nabla_{s\in S^c} Loss  )$\;
        \leavevmode\\
        $i = i + 1$\;
        \If{$pred \neq t$ and $i == max$}{
            $Population_s=$ \InitPopulation{$s_i, n$}\;
            $predictions=$ \GetPreds{$Population_s$}\;
            $elite=$ \FindElite{$predictions$}\;
            \If{$predictions[elite] < pred$}{
                $s_i=elite$\;
                $i = i - 100$\;
            }

        }
    }
    \KwRet $\hat{x}$\;
}
\end{algorithm}

\paragraph{Random Sampling} However, since DeepFake detections are used to detect whether an image is manipulated, most of the attribute variations can still be correctly classified by the detection. This leads to a large number of local optima or flat regions during adversarial optimization. 
Thus, the update process may fail to find the semantic perturbation that can successfully bypass DeepFake detections. 
To solve this problem, we propose to randomly sample style codes across various candidate attribute spaces to skip local optima or flat regions that can hinder traditional optimization methods. 
Specifically, we first sample three style codes with different variations in each candidate attribute space to form the candidate set of style codes. 
Then we generate a population of adversarial examples based on these style codes and feed them to the DeepFake detector to get predictions. 
Subsequently, we select the elite style code with lower fake predictions than the original and continue optimization with it.

\subsubsection{Black-box scenario} 
In the black-box scenario, the adversary has no knowledge of the DeepFake detector except for output. 
Thus, we cannot establish the loss function and optimize the semantic perturbations with the gradient information.
Considering the sensitivity of DeepFake detections to attribute manipulation, we utilize a genetic algorithm~\cite{bhandari1996genetic} to optimize semantic perturbations in the attribute space.
The genetic algorithm has a stronger global search ability and can better avoid falling into the local optimal solution.
It is formally described by Algorithm \ref{alg3}. 
In the following, we will introduce three steps in detail.


\SetKwRepeat{Do}{do}{while}%
\begin{algorithm}\footnotesize
\caption{Black-box Adversarial Optimization Function}
\label{alg3}
\KwIn{ A style code $s$; size of population $n$; target score threshold $c$; maximum iterations $max$; the DeepFake detector $D(\cdot)$; the semantic discriminator $D_s(\cdot)$; the face recognition model $F(\cdot)$.}

\SetKwFunction{Select}{Select}
\SetKwFunction{GetPreds}{GetPreds}
\SetKwFunction{FindElite}{FindElite}
\SetKwFunction{Selection}{Selection}
\SetKwFunction{Crossover}{Crossover}
\SetKwFunction{Mutate}{Mutate}

\SetKwFunction{Black}{Black-box Optimization}
\SetKwProg{Fn}{Function}{:}{}

\Fn{\Black{$s, c, n, max, D, D_s, F$}}{
    $Population_{s,0}=InitPopulation(s, n)$\; $i = 0$\;
    \While{$i < max$}{
        $Population_{s,i}=$ \Select{$Population_{s,i}, D_s, F$}\;
        $scores_{i}=$ \GetPreds{$Population_{s,i}, D$}\;
        $elite_{i}=$ \FindElite{$scores_{i}$}\;
        \If{$scores_{i}[elite_{i}] < c$}{
            $\hat{x} =G(elite_{i})$\;
            $break$\;
        }
        $parents_{i}=$ \Selection{$scores_{i}, Population_{s,i}$}\;
        $childs_{i}=$ \Crossover{$parents_{i}$}\;
        $childs_{i}=$ \Mutate{$childs_{i}$}\;
        $Population_{s, i+1}=childs_{i}\cup elite_{i}$\;
        $i = i + 1$\;
    }
    \KwRet $\hat{x}$\;
}
\end{algorithm}

\begin{itemize}[leftmargin=*]

\item \textbf{Step 1: Initialize Population}. We randomly sample $n$ style codes in various candidate attribute spaces to initialize the first population, namely $Population_{s,0}$. Then, these style codes are fed to the generator $G$ to generate corresponding adversarial examples.

\item \textbf{Step 2: Select Inconspicuous Individuals}. Since style codes are randomly sampled in the $S$ space, they may exceed the reasonable attribute space, resulting in abnormal adversarial examples. 
With this in mind, we first feed these style codes to the semantic discriminator $D_s$ to compute semantic loss. According to the loss, the style codes inside the attribute space will be selected. 
Meanwhile, we also feed these adversarial examples to the face recognition model $F$ to select the images with the same identity as the original image.

\item \textbf{Step 3: Generate New Population}. The selected adversarial examples are fed to the DeepFake detections to get their predictions and select the elite style code that obtains the smallest prediction score (the value represents the probability that the input image is a DeepFake image). According to the prediction scores, we can derive the next population by \textit{Selection}, \textit{Crossover}, and \textit{Mutation} operators described in \cite{bhandari1996genetic}. Furthermore, to prevent the optimal individual of the current population from being lost in the next generation, we employ the elitism strategy \cite{bhandari1996genetic} to add the elite in this population to the next.

\end{itemize}

\section{Evaluation}\label{section:evaluation}
In this section, we conduct comprehensive experiments to evaluate the attack performance of \system. 
Besides, we carry out a user study to evaluate the imperceptibility of the adversarial images generated by \system. 
Finally, we verify the effect of the latent code constraints and random sampling on \system's attack performance in the ablation study.

\subsection{Experiment Setup}
\paragraph{Datasets}
We consider two benchmark datasets, StyleGAN~\cite{karras2020analyzing} and FaceForensics++~\cite{rossler2019faceforensics++} to verify that our proposed attack is applicable to all types of DeepFakes.
\begin{itemize}[leftmargin=*]
\item \textbf{StyleGAN:} This dataset is the current state-of-the-art dataset for face synthesis. It is rich and diverse in age, race, and image background. We randomly select 1000 images in the StyleGAN dataset pre-trained on FFHQ~\cite{karras2019style} for evaluation.

\item \textbf{FaceForensics++:} This dataset is a large-scale dataset containing various types of DeepFakes, such as FaceSwap \cite{faceswap}, DeepFakes \cite{DeepFake}, Face2Face \cite{thies2016face2face}, and NeuralTextures \cite{thies2019deferred}. We extract images from videos in FaceForensics++ and randomly select 1000 images for evaluation.
\end{itemize}

\paragraph{Victim DeepFake Detections}\label{section:DeepFake detectors}
To demonstrate the effectiveness of \system in varying attack scenarios, we involve six white-box DeepFake detections developed by individual researchers and two black-box DeepFake detections developed by industry-leading vendors as our victim models. 
We describe them in detail as follows:
\begin{itemize}[leftmargin=*]
\item \textbf{White-box DeepFake detections:} To show the effectiveness of \system in bypassing arbitrary DeepFake detections, we choose those with different strategies and state-of-the-art performance from academic sources, as they are typically open-sourced. These DeepFake detections are XceptionNet~\cite{rossler2019faceforensics++}, FFD~\cite{dang2020detection}, Patch-forensics~\cite{chai2020makes}, CNNDetection~\cite{wang2020cnn}, Gram-Net~\cite{liu2020global}, and F$^3$-Net~\cite{frank2020leveraging}. 

\item \textbf{Black-box DeepFake detections:} 
In response to the growing DeepFake menace, a cohort of publicly available DeepFake detection APIs are designed by various vendors, e.g., Tencent~\cite{tencent} and Baidu~\cite{baidu}.
The model information of these APIs is confidential, with only the output being known. Thus, we select these commercial DeepFake detectors as our victim models in the black-box scenario.

\end{itemize}

\paragraph{Evaluation Metrics}
We utilize the following metrics for evaluating the effectiveness and imperceptibility of \system:
\begin{itemize}[leftmargin=*]
\item {\textbf{Accuracy}}: the ratio of images that are predicted as the original correct label by the DeepFake detections.
\item {\textbf{Attack Success Rate (ASR)}}: the ratio of adversarial examples that bypass DeepFake detections.
\item {\textbf{BRISQUE}}: a natural scene statistic-based distortion-generic blind/no-reference (NR) image quality assessment metric. We use BRISQUE instead of $L_p$ norms, PSNR, or SSIM to quantitatively evaluate the quality of adversarial examples. The main reasons are as follows: our attribute variation-based adversarial attack cannot be fairly evaluated by the $L_1$, $L_2$, $L_\infty$, PSNR, or SSIM metrics since the attribute variations vary greatly at the pixel level but have little impact on human perception. 
\end{itemize}

\paragraph{Attributes}
Following the annotation of forty binary attributes (with/without) in CelebA dataset~\cite{liu2015faceattributes} and combining the properties of the latent space of StyleGAN, eleven identity-preserving attributes are selected for our attack, including \textit{pale skin}, \textit{hair color}, \textit{hairstyle}, \textit{mouth open}, \textit{wearing lipstick}, \textit{bushy eyebrow}, \textit{eyebrow shape}, \textit{earring}, \textit{eyeball position}, \textit{eye close} and \textit{eyeglasses}, which cover most attributes that are not easily perceived by humans.

\subsection{Effectiveness against White-box DeepFake Detections}

\subsubsection{White-box DeepFake Detections}
Specifically, we involve six DeepFake detections that cover two major categories of DeepFake image detection in academia: Spatial-domain based detection and frequency-domain based detection.
For spatial-domain based detection, we select XceptionNet~\cite{rossler2019faceforensics++}, FFD~\cite{dang2020detection}, Patch-forensics~\cite{chai2020makes}, CNNDetection~\cite{wang2020cnn}, and Gram-Net~\cite{liu2020global}. They are trained on different network architectures and datasets. 
%
Besides, the processing strategies and focus of these methods are also different. For instance, CNNDetection incorporates various data augmentation strategies into the training process; Patch-forensics utilizes shallow networks with limited receptive fields to focus on small patches of the DeepFake image; Gram-Net focuses on the global image texture representations.
For frequency-domain based detection, we choose the representative and state-of-the-art DeepFake detection method, {F$^3$-Net}~\cite{frank2020leveraging}.
Note that since there is no official code and pre-trained model for F$^3$-Net, we train it with CelebA~\cite{liu2015faceattributes} and StyleGAN datasets following the unofficial implementation code \cite{f3net2020yike}.
The accuracy of these methods on two benchmark datasets is shown in Table \ref{tab:tab1}. 
Since the accuracy of CNNDetection, Gram-Net, and F$^3$-Net on the FaceForensics++ dataset is less than 50\%, these three are not selected for evaluation on the FaceForensics++ dataset.

\begin{table}
\centering
\caption{The accuracy of white-box DeepFake detections on two benchmark datasets. The \xmark refers to the accuracy is less than 50\%. }
\label{tab:tab1}
\begin{tblr}{
  cells = {c},
  hline{1-2,8} = {-}{},
}
\textbf{Mehtod} & \textbf{StyleGAN} & \textbf{\textbf{FaceForensics++}} \\
XceptionNet       & 86\%              & 93\%                              \\
FFD               & 94\%              & 94\%                              \\
Patch-forensics   & 91\%              & 96\%                              \\
CNNDetection      & 90\%              & \xmark                               \\
Gram-Net          & 100\%             & \xmark                               \\
F$^3-Net$         & 100\%             & \xmark                              
\end{tblr}
\end{table}

\subsubsection{Candidate Attributes}
In this part, we define two types of candidate attributes: Single candidate attributes and Multiple candidate attributes.
The Single candidate attributes refer to the adversary only selecting one attribute in eleven identity-preserving attributes as the candidate attributes of \system for each attack, which aims to evaluate the effectiveness of \system with different single attributes and explore the impact ranking of various attributes on \system's attack performance.
Furthermore, the Multiple candidate attributes refer to the adversary can select multiple attributes in eleven identity-preserving attributes to form the candidate attributes of \system. 
Here, we in turn add the attribute to the candidate attributes based on their impact ranking on \system's attack performance, from strongest to weakest. 
In this way, we can obtain ten kinds of Multiple candidate attributes.
Note that we also conduct the experiment with Multiple candidate attributes in other composition ways, including in turn adding the attribute to the candidate attributes based on their impact ranking on \system's attack performance, from weakest to strongest and the perceptual variation from small to large.
See more details in Appendix \ref{section:differentcombination}.

\subsubsection{Effectiveness}
In this part, we evaluate the effectiveness of \system with Single candidate attributes and Multiple candidate attributes respectively.

\begin{figure}[htp]
    \centering
    \includegraphics[width=\columnwidth]{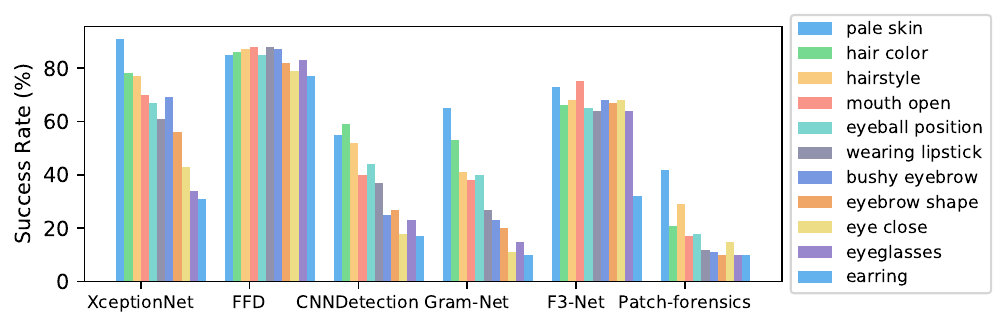}
    \caption{On the StyleGAN dataset, the attack success rate of \system with different Single candidate attributes against white-box DeepFake detections.}
    \label{fig:fig2}
\end{figure}

\begin{figure}[htp]
    \centering
    \includegraphics[width=8cm]{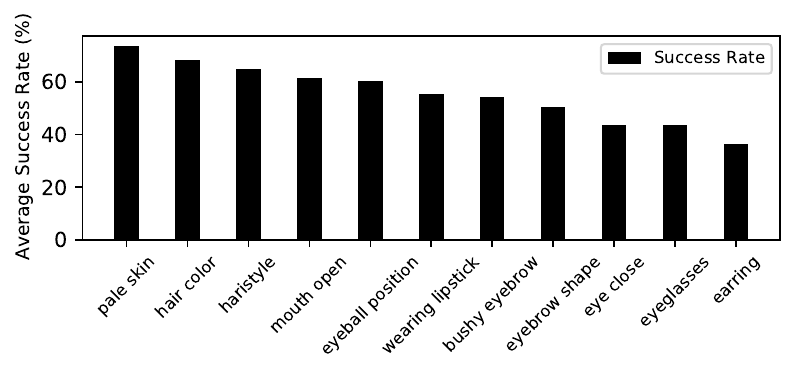}
    \caption{The impact ranking of various attributes on \system's attack performance.}
    \label{fig:fig3}
\end{figure}

\paragraph{Single candidate attributes}
Figure \ref{fig:fig2} shows the effectiveness of \system with different Single candidate attributes against white-box DeepFake detections. 
The experiment was carried out on the StyleGAN dataset.
We observe that \system with Single candidate attributes shows some level of effectiveness in attacking white-box DeepFake detections, but it is not considered satisfactory.
Besides, for the same DeepFake detection, the attack success rate of \system with different Single candidate attributes varies. 
For example, \system with the candidate attribute of \textit{pale skin} achieves a success rate of 91\% against XceptionNet, whereas \system with the candidate attribute of \textit{earring} merely a 31\% success.
However, from the overall trend, the impacts of various attributes on \system's performance are consistent across different DeepFake detections. 
We also carried out the experiment on the FaceForensics++ dataset and obtained the same conclusion. See details in Appendix~\ref{section:attributecombinationstrategy}.
Based on the above conclusions, we obtain the impact ranking of various attributes on \system's performance by calculating the average attack success rate, as shown in Figure \ref{fig:fig3}.
\vspace{-0.2cm}
\vspace{-0.2cm}

\begin{figure}[htp]
    \centering
    \includegraphics[width=8cm]{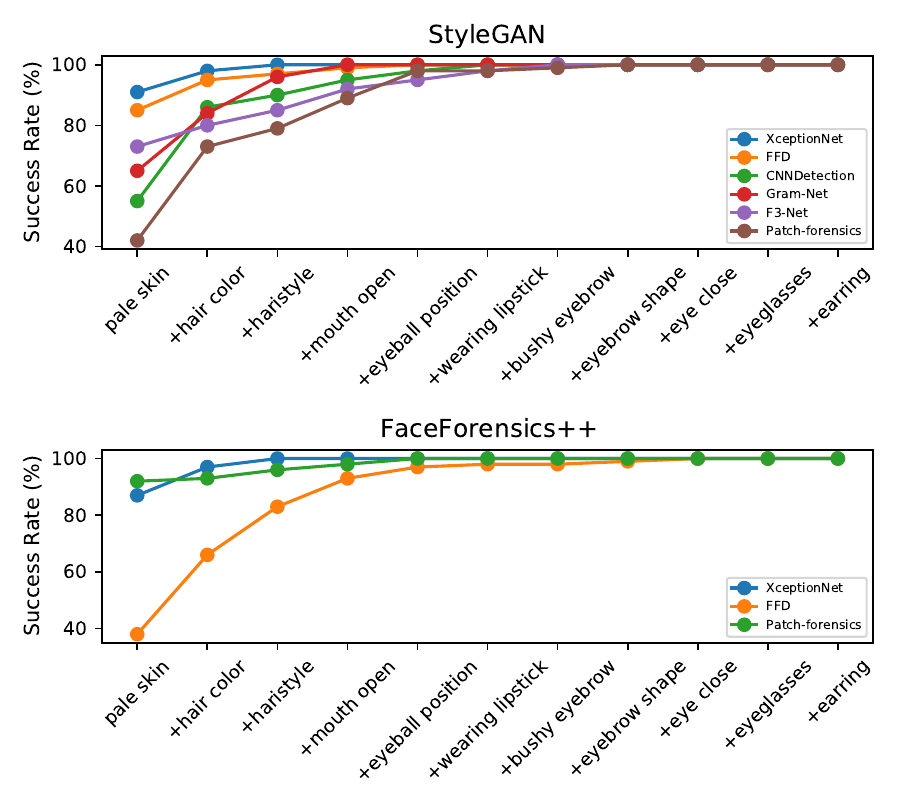}
    \caption{The success rate of \system with different Multiple candidate attributes against white-box DeepFake detections in two benchmark datasets. The "+" refers to adding the attribute to form the Multiple candidate attributes of \system.}
    \label{fig:fig4}
\end{figure}

\paragraph{Multiple candidate attributes}
Figure \ref{fig:fig4} shows the attack success rate of \system with different Multiple candidate attributes against white-box DeepFake detections. 
The results indicate that \system with over six candidate attributes can completely bypass all DeepFake detections involved in this study on two benchmark datasets. 
These findings suggest that the adversary can strategically choose more identity-preserving candidate attributes for the target person to achieve high bypass performance against DeepFake detections.
Besides, we observe that the effectiveness of \system is more prominent with an increase in the number of candidate attributes. 
This can be attributed to the fact that a larger number of candidate attributes expands the attribute space within which we search for semantic perturbations. Consequently, this expansion enhances the likelihood of discovering semantic perturbations that can successfully evade the DeepFake detections.

%
%

\subsubsection{Comparison with pixel-level-perturbation-based attacks}
To elaborate on \system's performance more convincingly, we compare \system attack extensively with three pixel-level-perturbation-based attacks, \textbf{FGSM}~\cite{goodfellow2014explaining}, \textbf{PGD}~\cite{madry2017towards}, and \textbf{CW2}~\cite{carlini2017towards}, which are frequently used in previous works~\cite{gandhi2020adversarial,carlini2020evading,neekhara2021adversarial,hussain2021adversarial,liao2021imperceptible,fan2021deepfake} to bypass DeepFake detection. For each attack, we evaluate both effectiveness and imperceptibility with a wide variety of different success rates by tuning their attack hyper-parameters. Note that, we adopt the BRISQUE metric to quantitatively evaluate the visual performance of the adversarial images here and a smaller BRISQUE score corresponds to more natural images.
For \system, we cover the success rate from low to high by controlling candidate attributes.

\begin{figure}[htp]
    \centering
    \includegraphics[width=8cm]{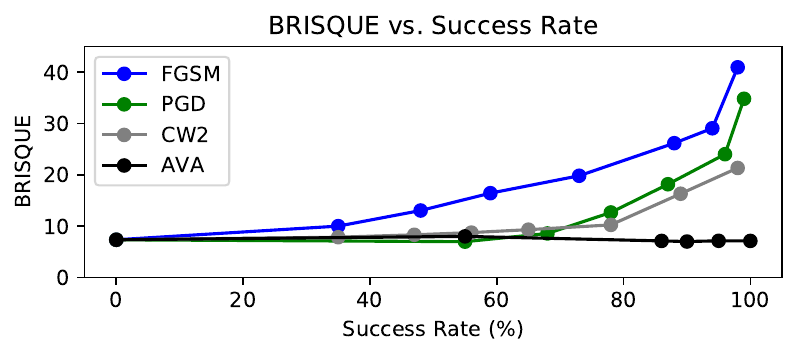}
    \caption{The Success Rate vs. BRISQUE of various attacks.  }
    \label{fig:fig5}
\end{figure}

\paragraph{Results} 
Figure \ref{fig:fig5} displays the average BRISQUE score of the adversarial examples generated by the four attacks under various attack success rates. 
The victim model is one of the state-of-the-art DeepFake detection methods, CNNDetection~\cite{wang2020cnn}, and the clean DeepFake images are from the StyleGAN dataset. 
We can observe that as perturbation increases, the success rate of the three pixel-level-perturbation-based attacks shows the same upward trend as \system. 
However, the imperceptibility of the adversarial examples generated by FGSM, PGD, and CW2 is progressively worse (i.e.\ ,looks more unnatural) as their attack success rate increases. 
while \system-generated adversarial examples remain at the same level as the original DeepFake images and even tend to improve. 
Therefore, we can conclude that \system can generate visually more inconspicuous and natural adversarial examples under a high attack success rate compared to the other three attacks, as it benefits from searching for semantic perturbation in the attribute space without introducing meaningless pixel-level perturbation noises.

\begin{table}
\centering
\caption{The accuracy of commercial DeepFake detection APIs on two benchmark datasets.}
\label{tab:tab2}
\begin{tblr}{
  cells = {c},
  hline{1-2,4} = {-}{},
}
\textbf{Method} & \textbf{StyleGAN} & \textbf{FaceForensics++} \\
Tencent                & 67\%              & 99\%                     \\
Baidu                & 100\%             & 95\%                     
\end{tblr}
\end{table}

\subsection{Effectiveness against Black-box DeepFake Detections}

\subsubsection{Black-box DeepFake Detections}
Existing commercial DeepFake detectors are mainly developed by industry-leading vendors, such as  Tencent~\cite{tencent}, Baidu~\cite{baidu}, Microsoft Video Authenticator~\cite{MicrosoftVideoAuthenticator}, RealAI~\cite{realai}, DuckduckGoose~\cite{duckduckgoose}, and RealityDefender~\cite{realitydefender}. Our threat model assumes that the adversary can only access the output of the black-box DeepFake detectors. These vendors enable developers to integrate their DeepFake detection techniques into their applications via APIs, regardless of their deep learning expertise. Consequently, we employ these APIs to obtain the required output.
Since we only get the DeepFake detection API support from Baidu and Tencent, we chose them as victim models in the black-box scenario. 
The accuracy of them on benchmark datasets is shown in Table \ref{tab:tab2}.

\subsubsection{Candidate Attributes}
In this part, due to the difficulty of the black-box scenario,
we choose the Multiple candidate attributes with eleven identity-preserving attributes.

\subsubsection{Effectiveness}
\begin{figure}[htp]
    \centering
    \includegraphics[width=8cm]{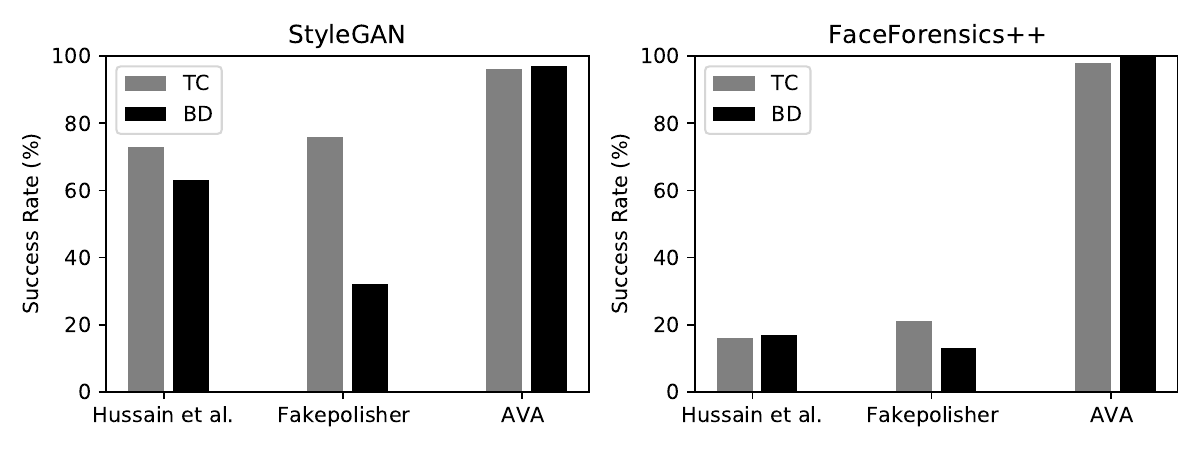}
    \caption{The attack success rate of various attacks against black-box DeepFake detections on two benchmark datasets.  }
    \label{fig:fig6}
\end{figure}

%
Figure \ref{fig:fig6} shows the effectiveness of \system against black-box DeepFake detections on two benchmark datasets. 
We can see that in the black-box scenario, our proposed \system attack achieves a success rate of over 96\% in bypassing all commercial DeepFake detection APIs on two benchmark datasets. 
These results indicate that our proposed \system attack poses a huge security risk to existing commercial DeepFake detection APIs, even though the adversary does not have access to their architecture and parameters. 
For instance, it is feasible to effortlessly release a DeepFake image, such as face swapping or face synthesis, onto downstream applications that integrate these APIs. In this way, the adversary can create a fraudulent account or defame the opponent candidate in a political campaign.

\subsubsection{Comparison with other black-box attacks} 
To elaborate on \system's effectiveness in the black-box scenario more convincingly, we compare \system attack extensively with other two 
state-of-the-art black-box attacks that are open-sourced, i.e., Hussain \textit{et al.}~\cite{hussain2021adversarial} and Fakepolisher~\cite{huang2020fakepolisher}. 
%
Hussain \textit{et al.} chooses a search distribution of random Gaussian noise around the original images, utilizes the Natural Evolutionary Strategies (NES)~\cite{wierstra2014natural, ilyas2018black} approach to estimate the gradient from queries, and then moves the original images in the direction of this gradient using iterative gradient sign~\cite{wang2021adversarial}. 
Fakepolisher~\cite{huang2020fakepolisher} is a reconstruction-based black-box attack, which performs shallow reconstruction for DeepFake images through a learned linear dictionary, intending to reduce the artifacts introduced during image synthesis.
For each attack, we follow the setting with its original paper.

\paragraph{Results}
Figure \ref{fig:fig6} shows the effectiveness of various black-box attacks against two commercial DeepFake detection APIs on the StyleGAN and FaceForensics++ datasets. 
Indeed, while Hussain \textit{et al.} and Fakepolisher have shown some ability to bypass certain commercial DeepFake detection APIs, their attack success rates remain below 20\% on the FaceForensics++ dataset, which does not meet our attack objectives.
%
In contrast, \system achieved a success rate of over 96\% in bypassing all commercial DeepFake detection APIs on two benchmark datasets. This success rate is significantly higher compared to other black-box attacks (p < 0.05).
%
%

\subsection{Imperceptibility to Humans}
Since the adversarial examples may not pose security risks if humans cannot perceive them as real and natural images, in this section, we conduct a human study to qualitatively evaluate the imperceptibility of adversarial examples generated by \system to humans. 

\paragraph{Human Study}
We set up a human study on Amazon Mechanical Turk (AMT) to ask human participants to evaluate the realness and naturalness of the given images. 
%
%
In the realness test, we would like to evaluate if an image is perceived as a DeepFake or not by humans. First, we designed the "\system \textit{vs.} Real" questionnaire with 50 pairs of images – one DeepFake image modified by \system and one real image from the FFHQ dataset. Then, we additionally designed the "DeepFake \textit{vs.} Real" questionnaire with 50 pairs of images – one clean DeepFake image from the StyleGAN dataset and one real image from the FFHQ dataset, to demonstrate that \system's modifications do not compromise the original authenticity of clean DeepFake images. 
Considering that some participants might not be familiar with DeepFake technology, we ask them to choose an image that is real. Note that, participants are not informed of how many images in each image pair are real, i.e., we designed a multiple choice question with 4 answers, "image 1 real", "image 2 real", "both real", and "both unreal";
In the naturalness test, we would like to demonstrate that the imperceptible semantic perturbations we modified on DeepFake images appear natural to humans.
Therefore, we designed the "DeepFake \textit{vs.} \system" questionnaire with 50 pairs of images - one clean DeepFake image and one DeepFake image modified by \system and asked participants to choose the image that they considered more natural. 
Note that the DeepFake images modified by \system from the one questionnaire are the same as the clean DeepFake images and correspond to the same real image. Therefore, to avoid the influence of similar image pairs on participants' decisions, we provide a separate set of 20 participants for each questionnaire, totaling 60 participants in our study.  
Additionally, we only recruit participants aged 18 and above, native English speakers from the United States, United Kingdom, and Australia with an answer approval rate of at least 95\%. We pay an average of \$22.00/hour per participant, which is higher than the average payout (\$11.00/hour) on the platform.
The Human Research Ethics Committee of the authors' affiliation determined that the study was exempt from further human subjects review, and we have followed the best practice for this ethical human subjects survey research.
Other details and examples are in Appendix~\ref{section:humanstudy}.

\begin{figure}[htp]
    \centering
    \includegraphics[width=8cm]{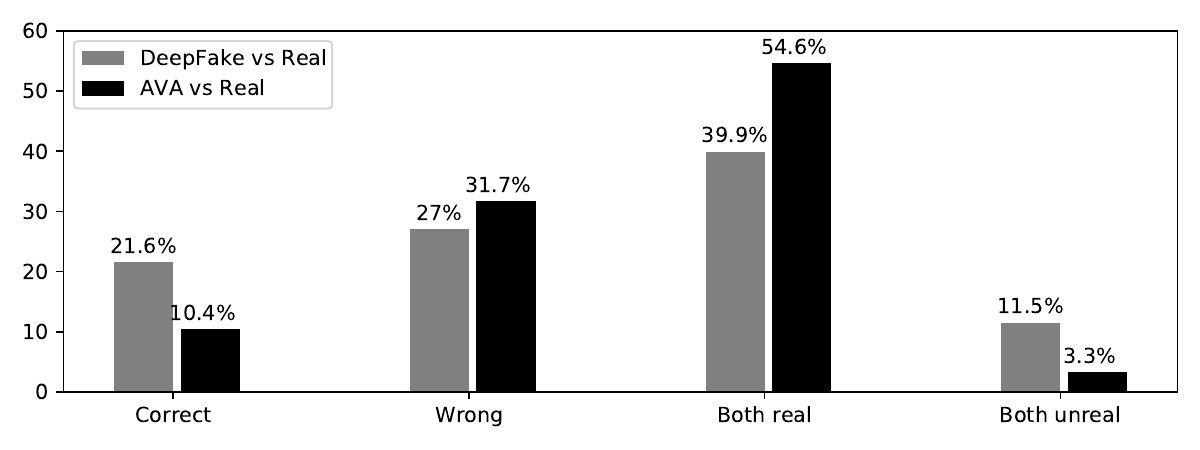}
    \caption{Results of the human study on evaluating realness. }
    \label{fig:fig7}
\end{figure}

\paragraph{Results}
Figure \ref{fig:fig7} shows the distribution of participants' responses on the realness test. 
Each legend entry represents a questionnaire. 
The x-axis in Figure \ref{fig:fig7} displays the four types of participants' responses. The term "Correct" indicates that participants correctly identify the adversarial examples. The term "Wrong" indicates that participants wrongly identify the adversarial examples as real. The term "Both real" indicates that participants consider the adversarial examples and the real images are both authentic. The term "Both unreal" is just the opposite. 
In the "\system \textit{vs.} Real" questionnaire, we can observe that over 50\% of the participants perceived the DeepFake image modified by \system to be just as authentic as the real images. This indicates that the DeepFake images modified by \system are realistic enough to fool human perception. Moreover, only 10.4\% of the participants correctly identified the DeepFake image modified by \system from real images, and 31.7\% of the participants gave exactly the opposite wrong answer. The accuracy of the participants is significantly lower than that of random guessing, which indicates that the DeepFake image modified by \system could not be identified by humans. 
Besides, the results of the "DeepFake \textit{vs.} Real" questionnaire demonstrate that clean DeepFake images have advanced to a level where they are difficult for humans to accurately distinguish. Moreover, our proposed \system's modifications do not compromise the original authenticity of DeepFake images.

In the "DeepFake \textit{vs.} \system" questionnaire, 49.75\% of participants selected the DeepFake images modified by \system as the more natural one, which is very close to the random guess (50\%). This indicates that the imperceptible semantic perturbations on DeepFake images appear natural to humans even when compared to the original DeepFake images.

\subsection{Ablation Study}
\subsubsection{The Effect of Latent Code Constraints and Random Sampling}
To further investigate the effect of latent code constraint described in section \ref{section:ganinversion} and \ref{section:adversarialattribute} and random sampling described in sections \ref{section:adversarialattribute} on \system's effectiveness and imperceptibility, we design three \system attacks with different settings for comparative analysis as follows: 
\begin{itemize}[leftmargin=*]
\item \textbf{AVA\_v0:} Without latent code constraints and random sampling.
\item \textbf{AVA\_v1:} With latent code constraints but without random sampling.
\item \textbf{AVA\_v2:} With latent code constraints and random sampling.
\end{itemize}
Then, on the StyleGAN dataset, we perform these three attacks with the candidate attribute \textit{pale skin} to bypass the DeepFake detection method, CNNDetection.


\begin{figure}[htp]
    \centering
    \includegraphics[width=8cm]{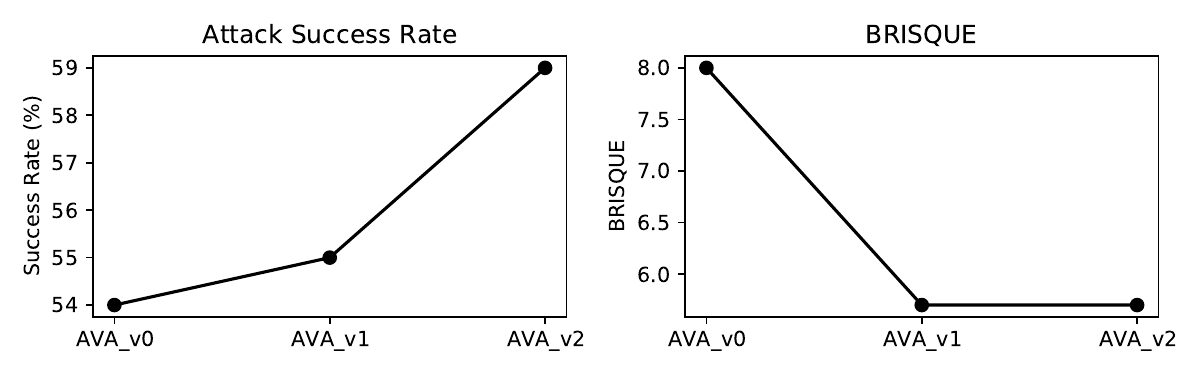}
    \caption{The effectiveness and imperceptibility of three different attacks. A smaller score of BRISQUE means the better perceptual quality. }
    \label{fig:fig8}
\end{figure}

\paragraph{Results} 
Figure \ref{fig:fig8} illustrates the success rate and BRISQUE score of the adversarial examples generated by these three attacks. The BRISQUE is used to quantitatively evaluate the imperceptibility of adversarial examples and a smaller score of BRISQUE means the better. 
We can observe that the adversarial examples generated by AVA\_v1 and AVA\_v2 achieve better visual performance than those generated by AVA\_v0. 
This demonstrates that constraining the latent code towards high-density portions of the StyleGAN's latent space and limiting the semantic perturbations within the reasonable attribute space can indeed improve the imperceptibility of the generated adversarial examples.

Moreover, the results of the attack success rate also show that AVA\_2 has better effectiveness than AVA\_0 and AVA\_1. 
This indeed provides evidence that random sampling across various candidate attribute spaces can effectively assist \system in skipping local optima or flat regions that can hinder traditional optimization methods.


\subsubsection{Why Impacts of Various Attributes on \system Vary}
\begin{figure}[htp]
    \centering
    \includegraphics[width=8cm]{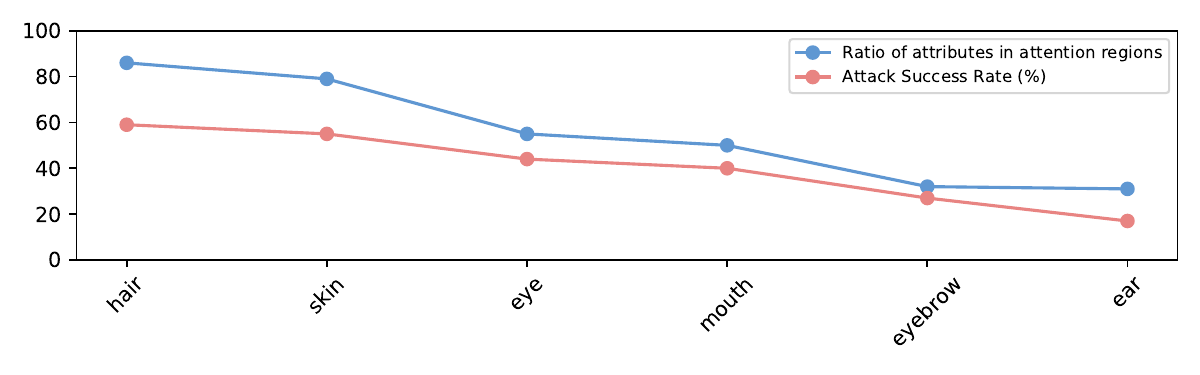}
    \caption{The ratio of various attributes appearing in CNNDetection's attention regions and the attack success rate of \system with various attributes against CNNDetection.  }
    \label{fig:fig9}
\end{figure}

Figure~\ref{fig:fig2} shows that the impacts of various attributes on \system's attack performance vary for the same DeepFake detection. 
We speculate that the main reason may be that DeepFake detection methods pay different attention to various attributes of the DeepFake images when detecting. Thus, perturbing the semantic perturbations on attributes that DeepFake detections pay more attention to is more likely to affect the output. 
To gain a deeper understanding, we select the DeepFake detection, CNNDetection~\cite{wang2020cnn} to analyze which attributes are more concerned during detection. 
Specifically, we randomly select 100 DeepFake images and utilize Grad-CAM~\cite{selvaraju2017grad} to localize the attention regions of CNNDetection. 
Then we manually count the ratio of various attributes appearing in CNNDetection's attention regions.
Figure \ref{fig:fig9} shows that the distribution of the ratio of various attributes appearing in CNNDetection's attention regions roughly matches the distribution of the attack success rate of \system with various attributes against CNNDetection, which verifies the correctness of the above assumption.
Of course, another explanation is also reasonable: the highest-ranking attributes were ones that tend to cover a large surface area (e.g., skin and hair color). Thus, making adjustments across a large part of the DeepFake image may remove more artifacts that would otherwise be detected.

\section{Countermeasures}\label{section:countermeasures}
In this section, we adopt two state-of-the-art defensive methods to evaluate the effectiveness of \system. Besides, we further discuss other possible mitigation strategies. 

\subsection{Effectiveness against Defenses}
\paragraph{Adversarial Training-based Defense}
We collect 20,000 real images from CelebA~\cite{liu2015faceattributes} and FFHQ~\cite{karras2019style} datasets and 20,000 DeepFake images from StyleGAN and ProGAN~\cite{karras2017progressive} datasets. Then, we adopt ResNet-50~\cite{he2016deep} pre-trained with ImageNet~\cite{deng2009imagenet} as our network architecture. Subsequently, we follow the training procedure from CNNDetection~\cite{wang2020cnn} to train the original DeepFake detector on the above datasets, denoted Detector$_{ori}$.
Then, we adopt the universal PGD-based adversarial training method~\cite{madry2017towards} as the adaptive defense that cannot access a set of \system-produced adversarial samples. 
In detail, we utilize a 10-step PGD L$_{\infty}$ attack with a step size of 1/255 and a perturbation size of 9/255 to train the robust DeepFake detector following the procedure of adversarial training in~\cite{madry2017towards}, denoted Detector$_{PGD}$. 
Similarly, following the procedure of adversarial training in~\cite{madry2017towards}, we also train a robust DeepFake detector using the adversarial examples generated by our proposed \system attack, denoted Detector$_{\system}$. 
Table~\ref{tab:tab3} shows the accuracy of various DeepFake detection methods on different datasets. 
We can observe that Detector$_{ori}$, Detector$_{PGD}$, and Detector$_{\system}$ all obtain high accuracy on clean DeepFake images. 
Meanwhile, both the PGD-generated and \system-generated adversarial examples against Detector$_{ori}$ effectively decrease the accuracy of Detector$_{ori}$ from 100\% to 0\%.   
However, these adversarial examples tend to lose their attacking effectiveness when confronted with robust DeepFake detection methods acquired through adversarial training.
For example, Detector$_{PGD}$ has 100\% accuracy on PGD-generated adversarial examples for bypassing Detector$_{ori}$ and Detector$_{\system}$ has 94\% accuracy on \system-generated adversarial examples for bypassing Detector$_{ori}$.

\begin{table}
\centering
\caption{The accuracy of various DeepFake detection methods on different datasets. The PGD$_{Detector_{ori}}$ refers to the PGD-generated adversarial examples against Detector$_{ori}$. The "-" means we don't need to get its result. }
\label{tab:tab3}
\begin{adjustbox}{width=\columnwidth,center}
\begin{tblr}{
  cells = {c},
  cell{3}{1} = {r=6}{},
  hline{1-3,9} = {-}{},
}
\textbf{Type}                     & \textbf{Dataset}   & \textbf{Detector\_ori} & \textbf{Detector\_PGD} & \textbf{Detector\_\system} \\
Clean DeepFake images & StyleGAN           & 100\%                  & 100\%                  & 93\%                   \\
Adversarial examples     & ${PGD_{Detector_{ori}}}$ & 0\%                    & 100\%                  & -                      \\
                         & ${\system_{Detector_{ori}}}$ & 0\%                    & -                      & 94\%                   \\
                         & ${PGD_{Detector_{PGD}}}$ & -                      & 91\%                   & -                      \\
                         & ${\system_{Detector_{PGD}}}$ & -                      & 0\%                    & -                      \\
                         & ${PGD_{Detector_{\system}}}$ & -                      & -                      & -                    \\
                         & ${\system_{Detector_{\system}}}$ & -                      & -                      & 0\%                    
\end{tblr}
\end{adjustbox}
\end{table}

Next, to better assess \system's effectiveness against the adaptive defense, we identify two distinct scenarios:
\begin{itemize}[leftmargin=*]
\item {\textbf{Defenders unknown to \system}}. In this scenario, we assume that defenders are unaware of the existence of our proposed \system attack. Therefore, defenders cannot utilize \system-generated adversarial examples to improve the robustness of DeepFake detection methods. 
We adopt the reliable PGD-based adversarial training method as the representative adaptive defense that cannot access a set of \system-produced adversarial samples. 
The Detector$_{PGD}$ shows a accuracy of 91\% against PGD-generated adversarial examples, indicating that the PGD-based adversarial training method significantly enhances the resistance of DeepFake detection to a wide range of adversarial attacks.
However, when it comes to our proposed \system attack, the accuracy of Detector$_{PGD}$ against \system-generated adversarial examples dropped to 0\% astonishingly. 
This suggests that \system reveals an unexplored area beyond previous pixel-level-perturbation-based attacks and poses a huge security threat to existing defenses.

\item {\textbf{Defenders known to \system}}. 
In this scenario, we assume that defenders are aware of the existence of \system. 
Therefore, defenders can utilize \system-generated adversarial examples to train the Detector$_{\system}$, a robust DeepFake detection method that is expected to defend \system. 
We use \system to attack Detector$_{\system}$ to evaluate \system's performance against a plausible adaptive defense that holds access to a set of attacker-produced adversarial samples. 
The result shown in Table~\ref{tab:tab3} indicates that our proposed \system can still totally bypass  Detector$_{\system}$. Of course, it requires more attack iterations than bypassing Detector$_{ori}$, but it is still acceptable to us. 
The main reason for the above phenomenon is that \system bypasses DeepFake detection methods by altering different facial attributes, which leads to introducing diverse perturbations. Therefore, it is hard for the defenders to comprehensively identify all possible perturbations due to the computational constraints.
\end{itemize}

\paragraph{Feature Squeezing-based Defense}
Feature Squeezing \cite{xu2017feature} detects adversarial examples by comparing a model's prediction on the original inputs with that on squeezed inputs, where the distance of this comparison is usually larger for adversarial examples than legitimate ones. We adopt the "Best Joint Detection" strategy in \cite{xu2017feature}, i.e., reducing the bit depth to 5 bits and using a 2 × 2 window for median filtering. 
Following the threshold selection strategy of the original paper, we use false positive rates of 5\%, 1\%, and 0\% for experiments, respectively. 
We randomly select 1000 original DeepFake images as legitimate examples and perform Feature Squeezing on them to obtain the squeezed examples. Then, we feed them to $Detector_{ori}$ and calculate the $L_1$ distance between the predicted score vectors of them. Finally, we combine the results and the false positive rate to choose the classification threshold. 

\begin{figure}[htp]
    \centering
    \includegraphics[width=8cm]{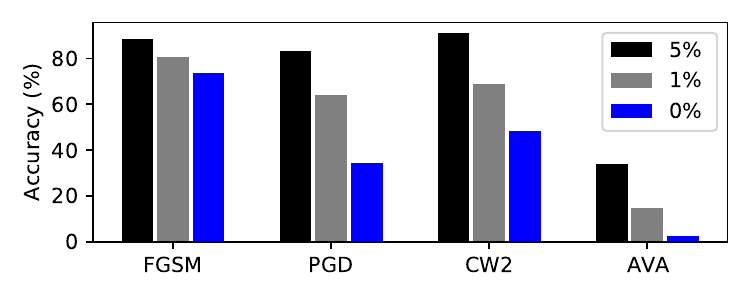}
    \caption{The accuracy of adversarial examples with Feature Squeezing under different false positive rates. }
    \label{fig:fig10}
\end{figure}
\vspace{-0.2cm}

Figure \ref{fig:fig10} elaborates on the accuracy of adversarial examples with Feature Squeezing under different false positive rates. We can see that after performing Feature Squeezing on the adversarial examples generated by FGSM, PGD, and CW2, more than 80\% of them are detected at the false positive rate of 5\%. In contrast, performing Feature Squeezing on the adversarial examples generated by \system only below 40\% of them can be detected at the false positive rate of 5\%. Moreover, we observe that Feature Squeeze is less effective to \system at the false positive rate of 0\%.

\subsection{Possible Mitigation Strategies}

Active defense strategies may be possible mitigation strategies to defend \system. The active defense strategies require humans to take some measures (e.g., adding crafted perturbations) before publishing images to the Internet to prevent or disturb DeepFake. In this way, not only our proposed \system attack cannot adversarially alter attributes to bypass DeepFake detection methods, but the abuse of DeepFake can be fundamentally solved. However, it is hard to identify the universal perturbations that are valid to all DeepFakes. Moreover, these measures, such as adding crafted perturbations or other noise, may fail when faced with denoising preprocessing or increased iterations during the GAN inversion module.

\section{Discussion}
\vspace{-0.2cm}

\paragraph{Ethical concerns and responsible disclosure} 
The Human Research Ethics Committee of the authors' affiliation determined that the study was exempt from further human subjects review, and we followed the best practices for ethical human subjects survey research. 
Besides, we conduct a comprehensive security evaluation on commercial detection detectors, which may raise some ethical concerns. Similar to the previous studies about the security of AI-powered systems~\cite{zhao2020large, shi2021adversarial}, we pay special attention to the legal and ethical boundaries. 
First, our evaluation of the commercial detection APIs strictly follows the official instructions. 
Besides, we have reported our results to the corresponding vendors and got their acknowledgments. We are now and will continue to engage with them to offer help with our best efforts. 

\paragraph{Limitations}
In the black-box scenario, we design a black-box \system attack based on the genetic algorithm. It requires a large number of queries to the victim model, which is not feasible for victim models with a limited number of queries. Moreover, the current black-box \system attack relies on the prediction scores returned by the victim model, which is not suitable for victim models that only return the final label category. We leave the in-depth exploration of more efficient black-box \system attacks for future work. 
Besides, due to the limitations of current attribute editing research, we cannot achieve precise control over more detailed attributes, such as fine hair. Therefore, AVA can be further enhanced based on the development of attribute editing.

\vspace{-0.2cm}

\section{Conclusion}
\vspace{-0.2cm}
In this paper, we have identified a new attribute-variation-based adversarial attack (\system) that could bypass DeepFake detections by perturbing the semantics in the attribute space of DeepFake images.
%
The empirical results demonstrate that our proposed \system attack can successfully bypass DNN-based DeepFake detections that dominate all DeepFake detection tools.
Moreover, \system also exhibits an effective penetrating capability in bypassing DeepFake detections enhanced with defenses, including adversarial example detection and adversarial training.
Meanwhile, \system achieves over 95\% success rate against commercial DeepFake detectors, which defeats the prominent state-of-the-art in black box attacks.
We also conducted a human study to evaluate the imperceptibility of the \system-generated adversarial examples and the results indicate that \system is often imperceptible to humans, which presents huge security and privacy concerns. 
%
%
Furthermore, we take the first step to investigate the vulnerability of DeepFake detection at the attribute level, hoping to shed light on directions to advance DeepFake detection.

\section*{Acknowledgement} We sincerely thank our shepherd and all anonymous reviewers for their constructive feedback.
This work was supported by National Natural Science Foundation of China under Grant No. 62372268, Shandong Provincial Natural Science Foundation (No. ZR2021LZH007, No. ZR2020MF055, No. ZR2022LZH013), Jinan City "20 New Universities" Funding Project (2021GXRC084).
Any opinions, findings, and conclusions in this paper are those of the authors and do not necessarily of supported organizations.

	\bibliographystyle{plain}
	\balance
	\bibliography{ava}
\appendices
\section{Effectiveness of \system with Single Candidate Attributes} \label{section:attributecombinationstrategy}

\begin{figure}[htp]
    \centering
    \includegraphics[width=\columnwidth]{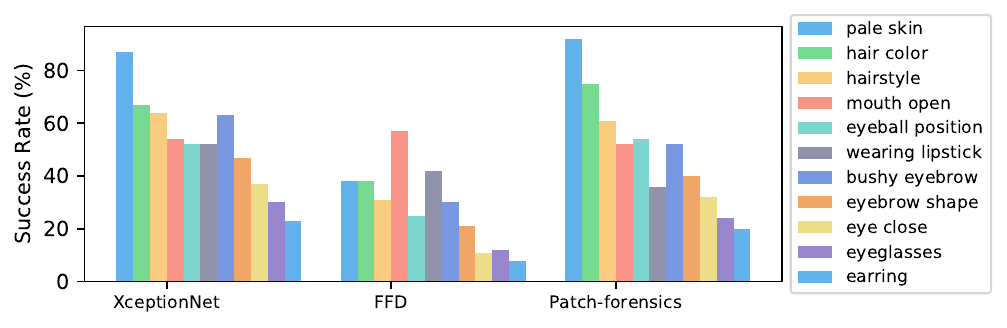}
    \caption{The bypass performance of \system attack with only one attribute against different victim models on the FaceForensics++ dataset.}
    \label{fig:fig11}
\end{figure}

We also carried out the experiment to evaluate the effectiveness of \system with different single candidate attributes against white-box DeepFake detections on the FaceForensics++ dataset. The results are shown in Figure \ref{fig:fig11}.

The results show the same trends as results on the StyleGAN dataset. Accordingly, we can determine that the impacts of various attributes on our attack are consistent across different model architectures and datasets. 

\section{Human Study}
\label{section:humanstudy}
\begin{figure}[h]
    \centering
    \includegraphics[width=\columnwidth]{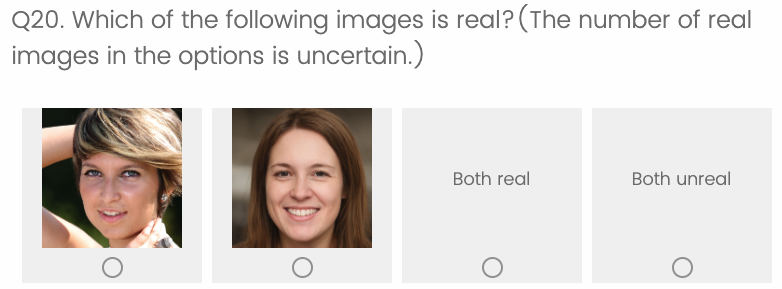}
    \caption{The typical example of "\system \textit{vs.} Real" questionnaire.}
    \label{fig:fig12}
\end{figure}

\begin{figure}[h]
    \centering
    \includegraphics[width=\columnwidth]{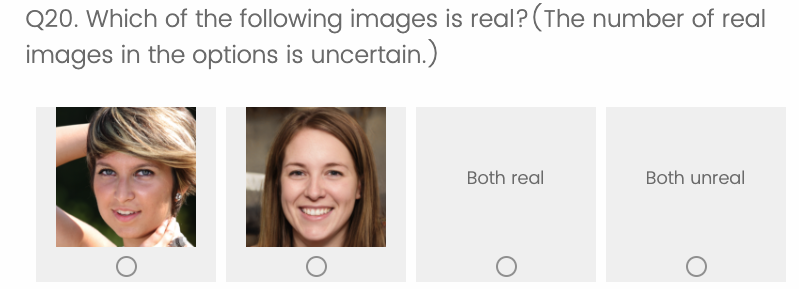}
    \caption{The typical example of "DeepFake \textit{vs.} Real" questionnaire.}
    \label{fig:fig13}
\end{figure}

\begin{figure}[h]
    \centering
    \includegraphics[width=\columnwidth]{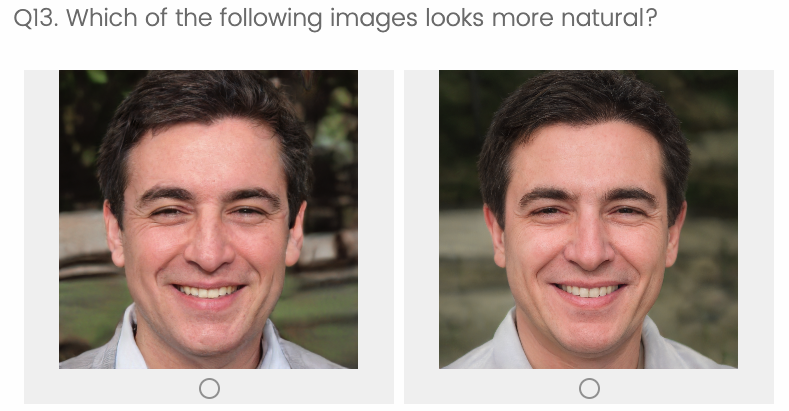}
    \caption{The typical example of "DeepFake \textit{vs.} \system" questionnaire.}
    \label{fig:fig14}
\end{figure}

We set up a human study on Amazon Mechanical Turk (AMT).
The AMT is a crowdsourcing platform to hire remotely located “crowd workers” to perform discrete on-demand tasks that computers are currently unable to do.
We follow a human study design from the literature \cite{zhang2016colorful}.
The order of these image pairs is randomized and participants are given unlimited time to respond. 
Furthermore, we would exclude participants who without good-faith effort (e.g., continuously if they pressed the same key on every trial). 
Figure~\ref{fig:fig12} shows an example of the "\system \textit{vs.} Real" questionnaire.
Figure~\ref{fig:fig13} shows an example of the "DeepFake \textit{vs.} Real" questionnaire.
Figure~\ref{fig:fig14} shows an example of the "DeepFake \textit{vs.} \system" questionnaire.
For complete information about all questionnaires please visit our anonymous website:\href{https://github.com/AnonymousUserA/AVA}{https://github.com/AnonymousUserA/AVA}.

\section{Attack Performance with Different Multiple Candidate Attributes}
\label{section:differentcombination}
we also conducted the experiment with multiple candidate attributes in other composition ways, including in turn adding the attribute to the candidate attributes based on their impact ranking on \system's attack performance, from weakest to strongest and the perceptual variation from small to large.

\begin{table*}[h]
\centering
\caption{The success rate of \system attack with multiple candidate attributes against six white-box DeepFake detection methods in two benchmark datasets. The "+" refers to adding the attribute to the multiple candidate attributes.}
\label{tab:tab4}
\scalebox{0.8}{
\begin{tblr}{
  column{2} = {c},
  column{3} = {c},
  column{4} = {c},
  column{5} = {c},
  column{6} = {c},
  column{9} = {c},
  column{10} = {c},
  cell{1}{1} = {r=2}{c},
  cell{1}{2} = {c=6}{},
  cell{1}{9} = {c=3}{},
  cell{3}{7} = {c},
  cell{3}{11} = {c},
  cell{4}{7} = {c},
  cell{4}{11} = {c},
  cell{5}{7} = {c},
  cell{5}{11} = {c},
  cell{6}{7} = {c},
  cell{6}{11} = {c},
  cell{7}{7} = {c},
  cell{7}{11} = {c},
  cell{8}{7} = {c},
  cell{8}{11} = {c},
  cell{9}{7} = {c},
  cell{9}{11} = {c},
  cell{10}{7} = {c},
  cell{10}{11} = {c},
  cell{11}{7} = {c},
  cell{11}{11} = {c},
  cell{12}{7} = {c},
  cell{12}{11} = {c},
  cell{13}{7} = {c},
  cell{13}{11} = {c},
  hline{1,3,14} = {-}{},
  hline{2} = {2-11}{},
}
\textbf{Attribute} & \textbf{StyleGAN } &               &               &               &               &                 &  & \textbf{FaceForensics++} &               &                 \\
                   & Xception           & FFD           & CNNDetection  & Gram-Net      & F$^3-Net$     & Patch-forensics &  & Xception                 & FFD           & Patch-forensics \\
earring            & 31\%               & 77\%          & 17\%          & 10\%          & 32\%          & 10\%            &  & 23\%                     & 8\%           & 20\%            \\
+eyeglasses        & 46\%               & 85\%          & 25\%          & 15\%          & 49\%          & 11\%            &  & 29\%                     & 16\%          & 29\%            \\
+eye close         & 68\%      & 87\%          & 34\%          & 25\%          & 71\%          & 15\%            &  & 42\%            & 23\%          & 42\%            \\
+eyebrow shape     & 84\%               & 91\%          & 50\%          & 46\% & 77\%          & 18\%            &  & 66\%                     & 57\%          & 62\%            \\
+bushy eyebrow     & 94\%               & 98\% & 63\%          & 70\%          & 86\%          & 24\%            &  & 90\%                     & 79\%          & 82\%            \\
+wearing lipstick  & 94\%               & 99\%          & 87\% & 90\%          & 89\%          & 32\%            &  & 97\%                     & 91\%          & 96\%            \\
+eyeball position  & 99\%               & 99\%          & 96\%          & 97\%          & 91\% & 50\%            &  & 100\%                    & 92\%          & 97\%            \\
+mouth open        & 100\%              & 100\%         & 99\%          & 99\%          & 97\%          & 69\%            &  & 100\%                    & 92\%          & 100\%           \\
+hairstyle         & 100\%              & 100\%         & 99\%          & 100\%         & 100\%         & 86\%            &  & 100\%                    & 96\% & 100\%           \\
+hair color        & 100\%              & 100\%         & 100\%         & 100\%         & 100\%         & 95\%            &  & 100\%                    & 97\%          & 100\%           \\
+pale skin         & 100\%              & 100\%         & 100\%         & 100\%         & 100\%         & 100\%           &  & 100\%                    & 100\%         & 100\%           
\end{tblr}}
\end{table*}

\begin{table*}[h]
\centering
\caption{The success rate of \system attack with multiple candidate attributes against six white-box DeepFake detection methods in two benchmark datasets. The "+" refers to adding the attribute to the multiple candidate attributes.}
\label{tab:tab7}
\scalebox{0.8}{
\begin{tblr}{
  column{2} = {c},
  column{3} = {c},
  column{4} = {c},
  column{5} = {c},
  column{6} = {c},
  column{9} = {c},
  column{10} = {c},
  cell{1}{1} = {r=2}{c},
  cell{1}{2} = {c=6}{},
  cell{1}{9} = {c=3}{},
  cell{3}{7} = {c},
  cell{3}{11} = {c},
  cell{4}{7} = {c},
  cell{4}{11} = {c},
  cell{5}{7} = {c},
  cell{5}{11} = {c},
  cell{6}{7} = {c},
  cell{6}{11} = {c},
  cell{7}{7} = {c},
  cell{7}{11} = {c},
  cell{8}{7} = {c},
  cell{8}{11} = {c},
  cell{9}{7} = {c},
  cell{9}{11} = {c},
  cell{10}{7} = {c},
  cell{10}{11} = {c},
  cell{11}{7} = {c},
  cell{11}{11} = {c},
  cell{12}{7} = {c},
  cell{12}{11} = {c},
  cell{13}{7} = {c},
  cell{13}{11} = {c},
  hline{1,3,14} = {-}{},
  hline{2} = {2-11}{},
}
\textbf{Attribute} & \textbf{StyleGAN } &       &              &          &           &                 &  & \textbf{FaceForensics++} &       &                 \\
                   & Xception           & FFD   & CNNDetection & Gram-Net & F$^3-Net$ & Patch-forensics &  & Xception                 & FFD   & Patch-forensics \\
~mouth open        & 70\%               & 88\%  & 40\%         & 38\%     & 75\%      & 17\%            &  & 54\%                     & 57\%  & 52\%            \\
+eye close         & 86\%               & 90\%  & 54\%         & 50\%     & 89\%      & 20\%            &  & 84\%                     & 75\%  & 63\%            \\
+eyeball position  & 96\%               & 90\%  & 84\%         & 90\%     & 93\%      & 31\%            &  & 94\%                     & 84\%  & 82\%            \\
+earring           & 97\%               & 91\%  & 85\%         & 91\%     & 93\%      & 34\%            &  & 99\%                     & 85\%  & 85\%            \\
+bushy eyebrow     & 99\%               & 91\%  & 93\%         & 94\%     & 95\%      & 39\%            &  & 100\%                    & 88\%  & 93\%            \\
+wearing lipstick  & 99\%               & 93\%  & 97\%         & 98\%     & 96\%      & 54\%            &  & 100\%                    & 91\%  & 96\%            \\
+eyeglasses        & 100\%              & 97\%  & 99\%         & 99\%     & 96\%      & 62\%            &  & 100\%                    & 92\%  & 98\%            \\
+eyebrow shape     & 100\%              & 100\% & 99\%         & 99\%     & 97\%      & 69\%            &  & 100\%                    & 94\%  & 100\%           \\
+hair color        & 100\%              & 100\% & 100\%        & 100\%    & 99\%      & 92\%            &  & 100\%                    & 95\%  & 100\%           \\
+hairstyle         & 100\%              & 100\% & 100\%        & 100\%    & 100\%     & 95\%            &  & 100\%                    & 98\%  & 100\%           \\
+pale skin         & 100\%              & 100\% & 100\%        & 100\%    & 100\%     & 100\%           &  & 100\%                    & 100\% & 100\%           
\end{tblr}}
\end{table*}

\newpage 

\section{Meta-Review}

The following meta-review was prepared by the program committee for the 2024
IEEE Symposium on Security and Privacy (S\&P) as part of the review process as
detailed in the call for papers.

\subsection{Summary}
The authors provide an approach to modifying deepfake images so that they can bypass deepfake detectors. The approach is based on modifying the semantics of an image (e.g., hair color), which has the effect of removing the artifacts that deepfake detectors use for detection. The authors provide a comprehensive analysis, demonstrating which semantics have the largest impact and the effectiveness when combined, using randomly sampled images from each different datasets. The authors have implemented a prototype to demonstrate the feasibility of the proposed attack and evaluate it on nine state-of-the-art DeepFake detection algorithms, including both white-box and black-box ones.

\subsection{Scientific Contributions}
\begin{itemize}
\item Provides a Valuable Step Forward in an Established Field
\item Identifies an Impactful Vulnerability
\end{itemize}

\subsection{Reasons for Acceptance}
\begin{enumerate}
\item The paper provides a valuable step forward compared to previous work since it demonstrates that is possible to circumvent existing DeepFake defenses, including black-box commercial systems.
\item The paper makes a comprehensive evaluation against nine state-of-the-art DeepFake detection algorithms and applications with a high success rate. The authors also built a prototype successful on white box and black box models.
\item The use of a GAN to control individual semantic features in the image independently is a novel idea, never been tried before.
\end{enumerate}


\end{document}